\documentclass[sigconf]{acmart}
\usepackage{booktabs} 

\usepackage{times}
\usepackage{latexsym}
\usepackage{graphicx}
\usepackage{url}
\usepackage{amsmath}
\usepackage{algorithmicx}
\usepackage[ruled]{algorithm}
\usepackage{algpseudocode}

\acmJournal{PACMHCI}
\acmVolume{9}
\acmNumber{4}
\acmArticle{39}
\acmYear{2010}
\acmMonth{3}
\acmArticleSeq{11}

\acmDOI{10.475/123_4}

\acmISBN{123-4567-24-567/08/06}

\begin{document}
\title{A Weakly Supervised Approach for Classifying Stance in Twitter Replies}


    

\author{Sumeet Kumar}
    \affiliation{
      \institution{Indian School of Business}
      \city{Hyderabad}
      \state{Telangana, India}
      \postcode{500032}
    }
    \email{sumeet_kumar@isb.edu}
\author{Ramon  Villa Cox}
    \affiliation{
      \department{School of  Computer Science}
      \institution{Carnegie Mellon University}
      \city{Pittsburgh}
      \state{Pennsylvania}
      \postcode{15213}
    }
    \email{rvillaco@cs.cmu.edu}

\author{Matthew Babcock}
    \affiliation{
      \department{School of  Computer Science}
      \institution{Carnegie Mellon University}
      \city{Pittsburgh}
      \state{Pennsylvania}
      \postcode{15213}
    }
    \email{mbabcock@cs.cmu.edu}

\author{Kathleen M. Carley}
    \affiliation{
      \department{School of  Computer Science}
      \institution{Carnegie Mellon University}
      \city{Pittsburgh}
      \state{Pennsylvania}
      \postcode{15213}
    }
    \email{kathleen.carley@cs.cmu.edu}






\renewcommand{\shortauthors}{Kumar et al.}

\begin{abstract}
Conversations on social media (SM) are increasingly being used to investigate social issues on the web, such as online harassment and rumor spread.  For such issues, a common thread of research uses adversarial reactions, e.g., replies pointing out factual inaccuracies in rumors. Though adversarial reactions are prevalent in online conversations, inferring those adverse views (or stance) from the text in replies is difficult and requires complex natural language processing (NLP) models. Moreover, conventional NLP models for stance mining need labeled data for supervised learning. Getting labeled conversations can itself be challenging as conversations can be on any topic, and topics change over time. These challenges make learning the stance a difficult NLP problem.

In this research, we first create a new stance dataset comprised of three different topics by labeling both users' opinions on the topics (as in pro/con) and users' stance while replying to others' posts (as in favor/oppose).  As we find limitations with supervised approaches, we propose a weakly-supervised approach to predict the stance in Twitter replies. Our novel method allows using a smaller number of hashtags to generate weak labels for Twitter replies. Compared to supervised learning, our method improves the mean F1-macro by 8\%  on the hand-labeled dataset without using any hand-labeled examples in the training set. We further show the applicability of our proposed method on COVID 19 related conversations on Twitter.

\end{abstract}

%
%



\maketitle

\section{Introduction}
Social Media analytics has been applied to many real-world problems. For example, one can estimate the popularity of a brand \cite{goh2013social}, understand the pattern in the diffusion of ideas \cite{pfeffer2014understanding}, and even predict the spread of diseases \cite{lee2013real}. These applications commonly use positive interactions, such as retweeting, liking, and following on Twitter. However, users can also convey negative opinions e.g., on Twitter, while replying to a tweet, a user can convey an opposing stance. Inferring stances in such replies, as in whether the reply post favors or opposes the original post, is increasingly being used in identifying misinformation \cite{zubiaga2015crowdsourcing} and online harassment \cite{li2016cyberbullying}, and  therefore, is an important research direction.

\begin{figure}[t]
    \centering
    \includegraphics[width=0.42\textwidth,height=0.22\textheight]{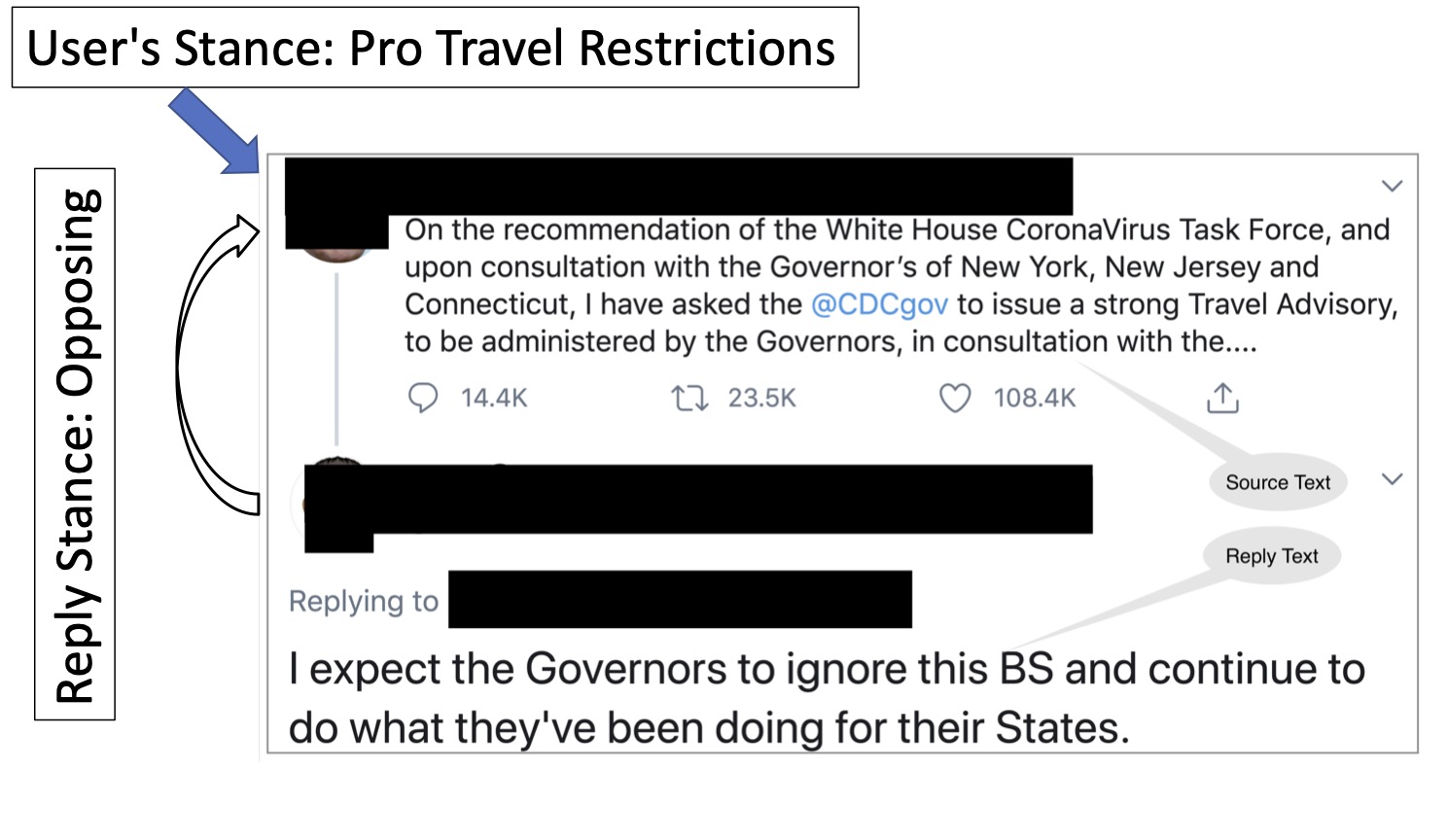}
    \caption{Twitter users exhibit their stance while in conversation with other users. In this illustrative example above, a user while replying to another user, reveals his stance by opposing to the original post.}
    \label{fig:reply_stance}
\end{figure}

To consider a concrete example, let's consider that a government agency (or a news agency) describes a new bill on Twitter (e.g., see Fig. \ref{fig:reply_stance}). Analyzing replies around the bill may allow a better estimate of the perceived public support. However, analyzing such conversations is difficult, and requires complex natural language processing (NLP) models. Moreover, conventional NLP models for analyzing conversations need labeled data for supervised learning. Getting labeled data is expensive as conversations can be about any topic, topics change over time, and new topics emerge, all of which makes learning the stance from conversations a challenging problem.

The problem that arises from topic-based conversations could be avoided by having a topic-agnostic model like a sentiment classifier \cite{liu2012sentiment,bing2014public}. However, sentiment classifiers are known to have a sub-optimal performance on stance classification tasks \cite{mohammad2017stance} as users can convey their favoring (or opposing) stance in a very sentiment neutral way. Therefore, better stance classifiers need to have contextual features of the target topic, e.g., a stance classifier on gun control needs to learn that `\#guncontrolnow' is a pro gun-control phrase, which a general sentiment classifier is unlikely to learn. 

One way to have topic relevant features is by having a training dataset on the topic of interest. Researchers have already created labeled datasets for a few controversial topics \cite{lu2015biaswatch,mohammad-etal-2016-semeval,constance}. However, labeled examples are few and are only available for a few topics, so training complex models on newer topics is still difficult. 

Instead of relying on labeled examples for training complex models, in this research, we propose a weakly-supervised approach to predict the stance  in Twitter replies. Using hashtags as weak labels for users' stance has been explored before \cite{zarrella2016mitre,lu2015biaswatch,misra2016nlds}, but using hashtags for learning stance in conversations is not straightforward. To the best of our knowledge, this is the first work that uses a few hashtags to first infer the stance of users and then use it to generate weak labels on Twitter replies. When compared with supervised learning, our proposed method improves the performance by 8\% (F1-macro) on a hand-labeled dataset without using any hand-labeled examples in the training set. To summarize, our main contributions are:


\begin{enumerate}
    \item To understand how Twitter users engage in conversations when  responding to other users' posts with similar of differing opinion, we create a new human labeled dataset with labels for both users' stance on a topic (as in pro/con) and the stance they take while replying (as in favor/oppose).
    \item  Using three different models, we show that predicting stance in Twitter replies using the traditional supervised approach is challenging. Therefore, we propose a weakly-supervised approach that uses a few labeled hashtags as initial seed labels. We show that using weakly labeled data, we are to able train large neural-network models that are data hungry, yet perform better than supervised models trained using leave-one-out event data. Prior work using weak supervision predicted users' stance on a topic (as in pro/con) and not stance in reply posts, which is our novel contribution. 
    \item We further show the applicability of our method using recent conversations on COVID 19. We validate our approach by finding groups of users who are pro/anti `Firing Dr. Fauci' and `Open American Economy' and by observing how these groups use different hashtags and websites.
\end{enumerate}

This paper is organized as follows. We first discuss the background and formulate the problem in Sec. \ref{sec:background}. We describe our new dataset in Sec. \ref{sec:dataset}  and then propose the methodology in Sec.  \ref{sec:methodology}. We discuss the experiments and results in Sec. \ref{sec:experiments} followed by an analysis of COVID-19 conversations on Twitter (Sec. \ref{sec:covid}). Related work which are not covered in introduction and background are described in Sec. \ref{sec:related_work}. Finally, we conclude and suggest directions for future research. 



\section{Background and Problem Formulation}
\label{sec:background}
Automated ways to learn stance -- which aim to predict the opinion of social media users on controversial topics -- has been broadly explored as two separate threads of research: 1) learning stance of users based on their social media posts about a topic \cite{mohammad2017stance,constance}, which we refer as users' stance, and  2) learning stance taken in conversations while replying, as in favoring or opposing  social-media posts \cite{zubiaga2016stance,kumar2019tree}, which we refer to as conversation or reply stance. Though these two approaches represent different aspects of stance-taking behaviour on social-media forums, it is natural to ask if they are two faces of the same phenomenon, and if so, what would be a unifying approach. 


In this research, our goal is to predict users' stance and stance in conversations. For this, we first use weak supervision to get the stance of a few seed users, and then use the seed users to learn the stance of the rest of the users in a Twitter network. As we would expect, given a discussion on a controversial topic, a user is more likely to oppose a post (while replying) of another user who has the opposing stance \cite{lai2019stance,derr2020link}.  Once stance of all users is known, we apply this idea to get weak labels for all conversations in the dataset. The weak labels are then used to train conversation classifiers which takes source texts, reply texts and weak labels as input.

\subsection{Problem Statement}
Given a topic, we retrieve a set of tweets $T=\{t_1, t_2, t_3, ...., t_m\}$ which were tweeted by a set of users $U= \{u_1, u_2, u_3,... u_n\}$ where $n \leq m$. Tweets are used to build user-hashtags network $H$ which is a weighted matrix created from $k$ most used hashtags. Similarly,  user-retweets is used to create a weighted matrix $R$ using $p$ most popular retweets in the dataset. Therefore, $H \in \mathbb{R}^{nxk}$ matrix and $R\in \mathbb{R}^{nxp}$ matrix.  $I$ is used to represent the union of $H$  and $R$.


An intermediate goal of this research is to correctly assign a stance label $\{+1, -1\}$ to as many users as possible in the set $U$ based on $T, H, R$ where ($+1$) indicates pro stance, ($-1$)  indicates anti-stance and ($0$) indicates unknown or neutral stance.  This assignment results in a user-stance matrix $S = \{s_1, s_2, s_3, ...., s_n\}$ where $s_i \in \{+1, 0, -1\}$. 

Let's define the conversations between two users $u_i$ and $u_j$ as $C_{ij}$ where $C_{ij} = \{C^1_{ij}, C^2_{ij},... C^k_{ij}\}$ are $k$ conversations (source tweet, reply tweets pair). Therefore, $C^k_{ij} = (t_s, t_r)$ where $t_s$ is source text and $t_r$ is reply text. These source reply pairs could be assigned stance labels as `Favoring' ($+1$)  and `Opposing' ($-1$) (or denying). Our proposed method tries to correctly assign a favoring/opposing stance label to as many conversations as possible.


\section{A New Stance Dataset}
\label{sec:dataset}

We create a new stance dataset that have stance labels for both users' stance and reply stance. The dataset is based on three recent events:  1) Student Marches (SM), 2) Santa Fe Shooting (SS), and 3) Iran Deal (ID). For each of the events, we collect tweets (including replies) and label a subset of the collected data. Due to space constraints, details on steps used to collect tweets is provided in Appendix at the end.

We describe the general statistics of the dataset in Tbl. \ref{tbl:contentious_topics1}, and then summarize the labeled part of the dataset in Tbl. \ref{tbl:chapter_6_dataset} and Tbl. \ref{tbl:labels_users}.


\begin{figure}[ht!]
    \centering

    \frame{\includegraphics[width=0.49\textwidth]{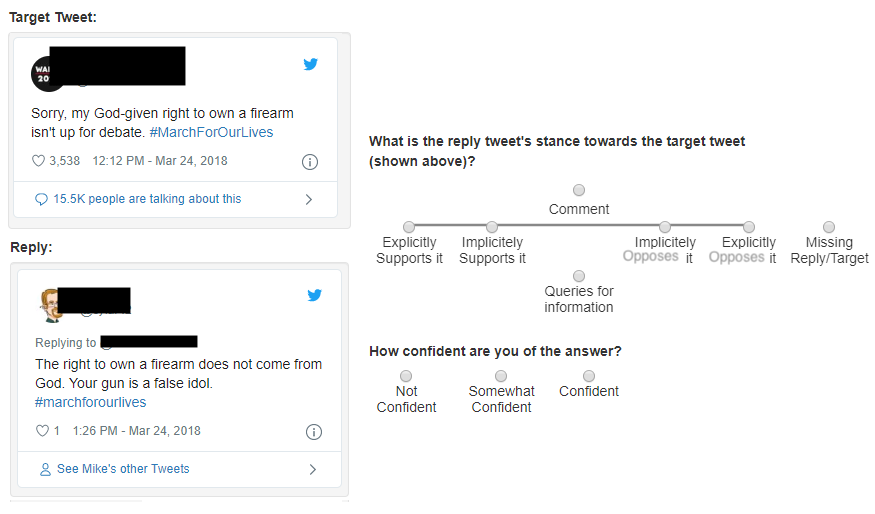}}
     \caption{Snapshot of the webpage developed for annotating replies. Annotators are required to provide the stance in the reply and their confidence in the provided label. }
      \label{fig:annex_pages}
\end{figure}


\begin{center}
\begin{table}[htb]
\small
\caption{Dataset summary}
\centering
 \begin{tabular}{|p{1.5cm}|p{.7cm}|p{.8cm}|p{.8cm}|p{.8cm}|p{.7cm}|p{.6cm}|} 
 \hline
 Events  & Users & Tweets & Retweets   & Hashtags & Replies \\ [0.5ex] 
 \hline
 \hline
Student Marches (SM)& 410785 & 1039778 & 391127& 17821 & 48277\\
 \hline
Santa Fe Shooting (SS)& 973506 & 3043731 & 910967& 45057 & 83293\\
 \hline
Iran Deal (ID)& 685058 & 3304519 & 580180& 86653 & 71808\\
 \hline
 \end{tabular}
 \label{tbl:contentious_topics1}
\end{table}
\end{center}

The annotation process was handled internally by research students in our group and for this purpose we developed a web interface for each type of response (see Fig. \ref{fig:annex_pages}). Each annotator was asked to go through a tutorial and a qualification test to participate in the the annotation exercise.

The annotator is required to indicate the stance of the response towards the target and also provide a level of confidence in the label provided. If the annotator was not confident in the label, then the task was passed to another annotator. If labels by both annotators agree, the label was accepted and if not the task was passed to a third annotator. Then the majority label was assigned to the response, and in the few cases were disagreement persisted, the process was continued with a different annotator until a majority label was found. Overall, 45\% of tweets were annotated only once, 47\% were annotated twice, 5\% were annotated three times and less than 2\% required more than three annotations.

\subsection{Definition of Classes}
We define the stance classes as:
\begin{enumerate}
    \item Explicit Oppose: Explicit Oppose means that the reply outright states that what the target tweets says is false or wrong.
    \item Implicit Oppose: Implicitly Oppose means that the reply implies that the tweeter believes that what the target tweet says is false/wrong/inaccurate.
    \item Implicitly Support: Implicitly Supports means that the tweet implies that the tweeter believes that what the target tweet says is true.
    \item Explicitly Support: Explicitly Supports means that the tweet outright states that what the target tweets says is true.
    \item Query: Indicates if the reply asks for additional information regarding the content presented in the target tweet.
    \item Comment: Indicates if the reply is neutral  regarding the content presented in the target tweet.
\end{enumerate}



As shown in Fig. \ref{fig:annex_pages}, annotators label twitter replies as one of the six classes, but for experiments the classes were later collapsed to four distinct classes (Support, Oppose Comment and, Query) by merging explicit and implicit categories. This led to the dataset as described in Tbl. \ref{tbl:chapter_6_dataset}. Note that Comment and Query represent neutral classes but we kept them separate as Query could also be used to question the content of original post, and hence opposing the original post.

 \begin{center}
\begin{table}[htb]
\centering
\small
\caption{Distribution of labeled replies}
 \begin{tabular}{|p{1.3cm}|p{1.8cm}|p{1.3cm}|p{1.1cm}|} 
 \hline
  Replies Stance   & Student Marches & Santa Fe Shooting & Iran Deal\\ [0.5ex] 
\hline
\hline
Oppose  & 220 & 304 & 198 \\
\hline
Support  & 212 & 225 & 202 \\
\hline
Comment & 293 & 246 & 153 \\
\hline
Queries & 42 & 21 & 19 \\
\hline
\end{tabular}
\label{tbl:chapter_6_dataset}
\end{table}
\end{center}


\subsection{Inter Annotator Agreement}

To validate the methodology, we selected 55\% of the tweets that were initially confidently labeled to be annotated again by a different team member. Of this sample, 86.83\% of the tweets matched the original label and the remainder required additional annotation to find a majority consensus. From the 13.17\% of inconsistent tweets, a 61.86\% were labeled confidently by the second annotator. This means that among the confident labels we validated, only 8.15\% resulted in inconsistencies between two confident annotators, which we deemed an acceptable error margin.

Cohen’s kappa measures the agreement between two or more raters. If each rate labels N items into C categories, Cohen kappa is defined as:

\begin{equation}
\kappa = \frac{p_0 - p_e}{ 1 - p_e}     \\
        = \frac{0.92 - 0.33}{ 1 - 0.33} \\
        = 0.89
\end{equation}

where $p_0$ is the relative observed agreement among raters and $p_e$ is the estimate of possible agreement by chance. In our experiment, $p_0 = 0.92$ and agreement chance $p_e = 0.33$ as there are three class types. This leads to $\kappa$ value of 0.89

\subsection{Users' Stance}

In addition to conversations, some users in the dataset are also labeled for their stance. For `Santa Fe Shooting' and `Student Marches', the stance was labeled for `pro/con' gun control. For `Iran Deal', the stance was evaluated for pro and against the breaking of the Iran deal agreement. The labeling exercise involved reading all tweets of selected users in the dataset and suggesting a final label (pro/con) based on the overall opinion. For most users, the number of tweets were in the range of 1 to 20 where the larger number of tweets mostly coming from public figures and news agencies.  Using multiple tweets was useful as we find that often one tweet on a topic is not sufficient to get the correct label. Especially certain news agencies tried to portray neutral but their inclination becomes clear by  observing their multiple tweets. The final labeled dataset is summarized in Tbl. \ref{tbl:labels_users}.

\begin{center}
\begin{table}[htb!]
\caption[Distribution of labeled users' stance]{Distribution of labeled users' stance.}
\centering
\small
\begin{tabular}{|p{1.3cm}|p{1.8cm}|p{1.3cm}|p{1.1cm}|} 
 \hline
  Users' Stance   & Student Marches & Santa Fe Shooting & Iran Deal\\ [0.5ex] 
\hline
\hline
Pro  & 129 & 188 & 137 \\
\hline
Anti  & 154 & 64 & 122 \\
\hline

\end{tabular}

\label{tbl:labels_users}
\end{table}
\end{center}

\begin{figure*}[htb!]
    \centering
    \includegraphics[width=0.69\textwidth]{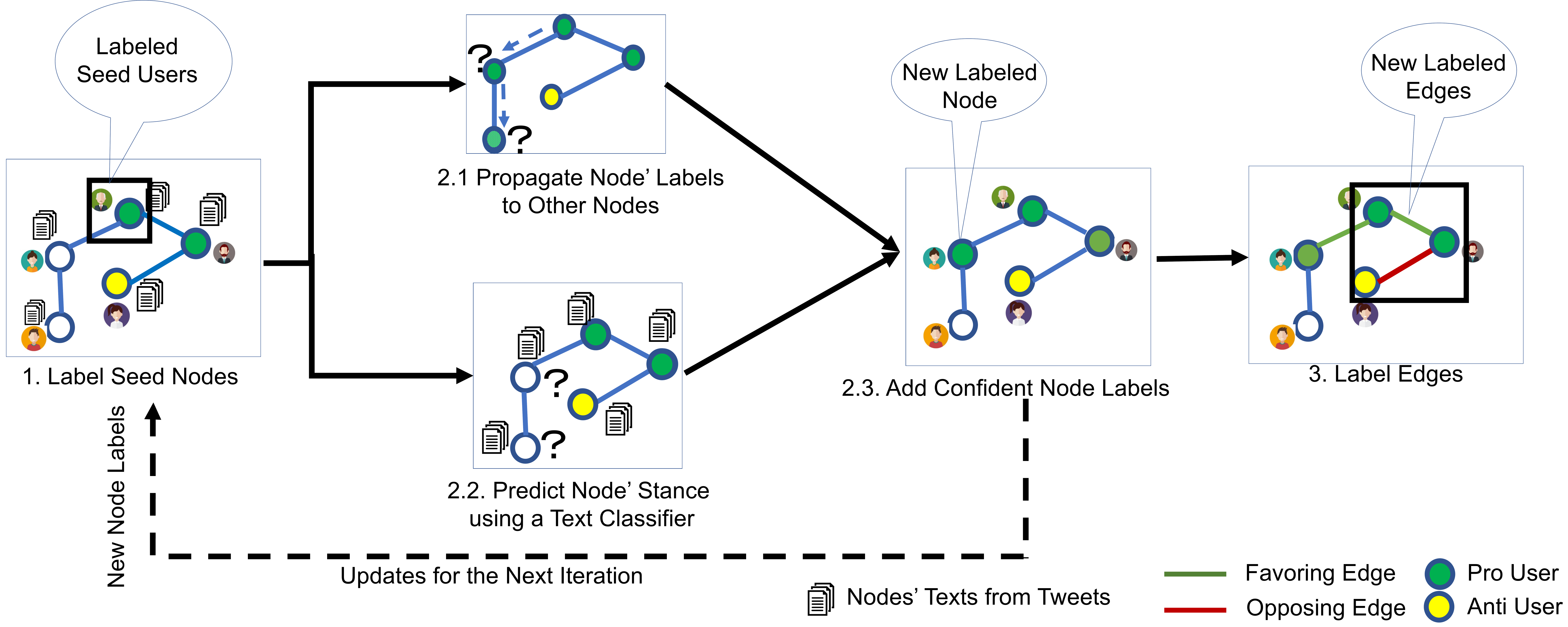} 
    \caption[System Design]{A high level illustration of the proposed system. The entire methodology could be divided in three steps. In the first step, we set seed user labels using labeled hashtags, the second step is an iterative co-training process, and the third step generates weak labels for edges that represent user-to-user conversations.}
    \label{fig:cotrain_model_digram}
\end{figure*}

\section{Methodology}
\label{sec:methodology}



We propose a three step process (see Fig. \ref{fig:cotrain_model_digram}). we first use weak supervision to get the stance of a few seed users (Step 1), and then use the seed users to learn the stance of the rest of the users' (Step 2). Then we use users' stance to get weak labels for stance in all conversations between the users in the dataset (step 3). The labeled  conversations are then used to train conversations based stance classifiers.


\subsection{Step 1: Label Seed Users by Labeling Hashtags}
We use two to six hashtags as the weak signal to label a few seed users. This step uses the label propagation algorithm on the user-hashtag bipartite network. 

\begin{equation}
    S^I =  H \cdot \tilde{h}
\end{equation}

where $S^I$ is the resulting labeled  users and $\tilde{h}$ indicates the hashtags vector with a few labeled hashtags as $\{ +1, -1\}$. How the hashtags are picked for labeling is described in experiments.

\subsection{Step 2: Propagate Seed Node Stance Labels to Other Nodes using Co-Training}
We observed that a graph based classification method on hashtags or retweets network has high precision but low recall for users' stance prediction. This is because not all users use hashtags or retweets resulting in only a partial fraction of users getting labeled by the graph based methods using data $I$. To resolve, researchers have used various approaches, e.g., \cite{lu2015biaswatch} used similarity between users based on their text usage to supplement the original retweets network. A similarity based method ($N^2$) does not scale well on large networks as a comparison between any two nodes is needed. Therefore, we use a co-training based approach which again uses users' text features and users' networks but is more scalable and has shown to improve the classifiers performance in different problem domains \cite{yu2019co,liu2013adaptive,mo2010exploit}.

We propose co-training two classifiers, 1) a bi-partite label propagation on $I$, and  2) a text classifier that uses text messages posted by users'. A final joint model that uses predictions of the two classifiers is used to get the final users stance.

\subsubsection{Bi-Paritie Network Label Propagation}
A bi-partied label propagation algorithm is used to propagate labels to other nodes. For a bipartite network $I$, the stance propagates in two steps: 1) label propagates from users to other node types (hashtags or retweets), and in step 2) label spreads from other nodes to users. $\theta_u$ to be a parameter that acts as a threshold for spreading the labels from other node types to users. A label propagation model could be described as:


\begin{eqnarray}
    \tilde{S} \leftarrow \sigma_{\theta_h}{(I' \cdot S^I)} \\
    S^I \leftarrow  \sigma_{\theta_u}{(I \cdot \tilde{S})}
\end{eqnarray}

where $\cdot$ is dot product, and $I'$ is the transpose of the matrix $I$. For influence function, we use $\sigma$ as used in Linear Threshold Model (LTM) and allows propagating the stance to only the strongly connected neighbours \cite{chen2010scalable}. An LTM model is described as:


\begin{equation}
   \sigma_{{\theta_u}_{j}} =\left\{
      \begin{array}{@{}ll@{}}
        1, & \text{if}\ \sum_{k=1}^n I_{jk} * \tilde{S}_k  > \theta_u \\
        -1, & \text{if}\ \sum_{k=1}^n I_{jk} * \tilde{S}_k  < -\theta_u \\
        0, & \text{otherwise} \\
      \end{array}\right.
\end{equation}
where  $s^I$ is users' stance based on network data. 

We also define a confidence estimate in predicting users stance which is defined as the ratio of weight of the edges leading to the stance of a user, divided by the sum of all edge-weights for that user:

\begin{equation}
  c^I_j=\left\{
      \begin{array}{@{}ll@{}}
         \frac{\sum_{k=1}^n I_{jk} * \mathbf{1}_{\tilde{s}_k = s^I_j}}{\sum_{k=1}^n I_{jk} }, \text{if}\ s^I_j  \neq 0 \\
         0, & \text{otherwise} \\
      \end{array}\right.
\end{equation}

where $c^I_j$ is the confidence in estimating the stance $s_j$ of user $j$. 

\subsubsection{Text Based Users' Stance Classifier}
In this step, we use users' text data (from their tweets) to train an SVM text classifier. The text classifier is than used to derive the stance of users. Though the text classifier only predicts one label for a text sample (obtained from a tweet), because users can have multiple tweets, multiple predictions (one for the text from each tweet) can be used to quantify the confidence in the stance estimation of a user. Let's define the text classifier as $f_{text}(\theta, t)$ where $\theta$ is the training parameter and $t$ is tweet's text. Based on the predictions of the trained model $f_{text}(\theta, t)$, the number of `pro' text and `con' text a user has, determines the stance $S^T$ and prediction confidence $c^T$. This step can be summarized as:

\begin{equation}
   S^T_j=\left\{
      \begin{array}{@{}ll@{}}
        1, & \text{if}\ \frac{\sum_{k=1}^m \mathbf{1}_{f_{text}(\theta, t_k) > 0}}{\sum_{k=1}^m  1}  > \theta^T  \\
        -1, & \text{elif}\ \frac{\sum_{k=1}^m \mathbf{1}_{f_{text}(\theta, t_k)< 0}}{\sum_{k=1}^m  1}  > \theta^T \\
        0, & \text{otherwise} \\
      \end{array}\right.
      \label{eqn:text1}
\end{equation}

\begin{equation}
   c^T_j=\left\{
      \begin{array}{@{}ll@{}}
        \frac{\sum_{k=1}^m \mathbf{1}_{f_{text}(\theta, t_k) > 0}}{\sum_{k=1}^m  1}, \text{if}\   s^T_j > 0  \\
        \frac{\sum_{k=1}^m \mathbf{1}_{f_{text}(\theta, t_k) < 0}}{\sum_{k=1}^m  1}, \text{elif}\   s^T_j < 0  \\
        0, & \text{otherwise} \\
      \end{array}\right.
      \label{eqn:text2}
\end{equation}

where $s_k$ is the stance of the $k^{th}$ tweet of the user $j$ who has a total of $m$ tweets, and $\theta^T$ is a linear threshold parameter determined using experiments. Like in the previous step, we only use the users who pass the threshold criterion and for whom stance is not zero in co-training iterations as described next.

\subsubsection{Co-training Users' Text Classifier and Network Label Propagation}
The steps for co-training of text classifiers is described in Alg. \ref{alg:co_train}.

\begin{algorithm}[htb!]
  \caption{Co-Training of network based label-propagation model and text classifier
    \label{alg:co_train}}
  \begin{algorithmic}[0]
    \Require{$T$ is the set of tweets collected}
    \State \Function{Co-Train  Classifiers}{$T$}
        \State Extract User-Text D 
        \State Extract User-Hashtag Network H
        \State Label Seed hashtags \Comment{e.g, \#Prochoice +1}
        \State Get Seed users (UL)  
        \State Get Unlabeled users (UU)  
      \While{until convergence}
      \State STEP 1: Label Propagation 
          \State Label nodes of I using UL
          \State Predict the Stance $S^I$ of UU using I
          \State Estimate stance confidence ($c^I$)
          \State STEP 2: Text based  classification 
          \State  Train a text classifier $f_{text}(\theta)$ using UL
          \State Use $f_{text}(\theta)$ to Predict UU stance $S^T$
          \State Estimate stance confidence ($c^T$) 
          \State STEP 3:  Update UL 
          \State L=AddNewLabeledExamples($S^I, c^I,S^T,c^T$) 
          \State UL= UL + L
      \EndWhile
      
      \State \Return{$UL$}
    \EndFunction
  \end{algorithmic}
\end{algorithm}

At the end of a co-training iteration, we add new more confident labeled examples to expand the training set \cite{blum1998combining}. For `AddNewLabeledExamples',  we use the $K$ percentage top confident labels of both classifiers to be added as new training examples where $K$ is a hyper-parameter with range 2\% to 20\%. Its value is decided using experiments. 

\subsubsection{Joint Users' Stance Model}
For the final users' stance prediction, we create a joint model that combines the predictions of the two models using the confidence scores to create a classifier (as described below).

\begin{equation}
      s_j=\left\{
      \begin{array}{@{}ll@{}}
            0, \text{if}\ c^T_j  = c^I_j = 0\\
         S^T_j , \text{elif}\ c^T_j  >=  c^I_j \\
         S^I_j, \text{Otherwise}
        
      \end{array}\right.
\end{equation}


\subsection{Step 3: Generate Weak Labels for Conversations from Users' Stance}
\begin{figure}[htb!]
    \centering
    \includegraphics[width=0.48\textwidth]{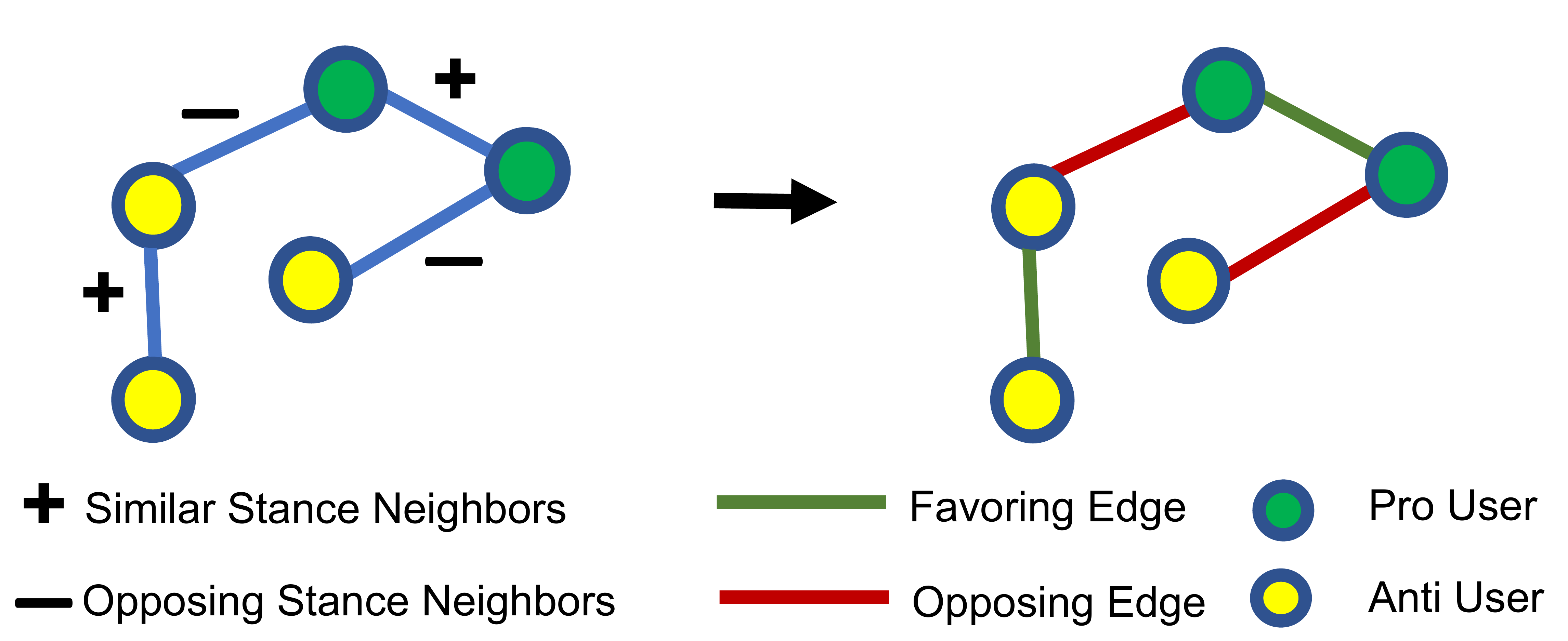}
    \caption{Model of predicting stance of edges (conversations) as in favoring/opposing from stance of users (pro/anti)}
    \label{fig:edge_stance_from_users_stance}
\end{figure}

 User's characteristics can be used to predict signs in networks \cite{yang2012friend}.  If two users have the same stance (i.e. either both of them are pro or both of them are anti), the conversations between them (as in edges in the network) are more likely to be `favoring' (see Fig. \ref{fig:edge_stance_from_users_stance}). In contrast, if two users have different stances, the conversations between them are expected to be `opposing'  \cite{lai2019stance}. 
 

This way, we can get weak edge labels for any conversations (source, reply pair text) between any two users (see Eqn. \ref{eqn:conv1}).  As described earlier, we use co-trained models to predict the stance of as many users as possible in the network, which in-turn leads to stance for most conversations between users.


\begin{equation}
   C^k_{ij}=\left\{
      \begin{array}{@{}ll@{}}
        0, & \text{if}\  s_i = 0 \  or \  s_j = 0 \\
        -1, & \text{elif}\  s_i \neq s_j\\
        1, & \text{otherwise} \\
      \end{array}\right.
      \label{eqn:conv1}
\end{equation}





\subsection{Step 4: Train Conversation Stance Classifier using Weak Labels }
The conversations based text classifiers use conversations (source, reply pair) between any two users that are labeled (last step).  The classifiers are trained on this source-reply-label data, and are then used to predict the stance labels for the rest of conversations (or the test set for validation). The choice of classifiers and the various alternatives are described in the experiments section.

To summarize the steps, we propose an approach to train a conversation (source text, reply text) based stance classifier, with only weak supervision from a few labeled hashtags. Note that we still use labeled examples to quantify the performance of the models.

\section{Experiments and Results}
\label{sec:experiments}

\subsection{Seed Hashtags and Seed Users}
\label{sec:seed_tags}

A small fraction of data points (seed users) are needed in the begining. To get seed user labels, we manually label a few (two to six) popular hashtags which we call as seed hashtags (details in Tab. \ref{tbl:seed_hashtags2}) which give clear stance signal.  The labels given by these seed-hashtags are propagated to users using the co-training algorithm as described earlier (Step 2). The final output of co-training is the stance labels for all users, which are then used to determine stance labels for all conversations. The labeled conversations are then used for training classifiers.
 




\begin{center}
\begin{table}[htb!]
\centering
\small
\caption{Seed hashtags and their labels}
 \begin{tabular}{|p{1.7cm}|p{4.3cm}|} 
 \hline
 Dataset  & Seed Hashtags \\ [0.5ex] 
 \hline
 \hline
  Iran Deal & \#thankyoutrump: Pro, \#iranuprising: Anti, \#freeiran: Anti \\
 \hline
 Student Marches &   \#defendthesecond: Pro, \#2ashallnotbeinfringed: Pro,\#2adefenders: Pro, \#2ndamendment: Pro ,\#guncontrolnow: Anti, \#marchforourlives: Anti  \\
 \hline
 Santa Fe Shooting & \#defendthesecond: Pro, \#2ashallnotbeinfringed: Pro,\#2adefenders: Pro, \#2ndamendment: Pro ,\#guncontrolnow: Anti, \#marchforourlives: Anti  \\
 \hline
 \end{tabular}
 \label{tbl:seed_hashtags2}
\end{table}
\end{center}


\subsection{Text Pre-Processing}
As we have sentence pairs (source, reply) as input, we use features extracted from text to train the models. For each sentence pair, we extract text features from both the source and the reply separately. Before using the text, we perform some basic text cleaning which involves removing any @mentions, any URLs and the Twitter artifact (like `RT') which gets added before a re-tweet. Some tweets, after cleaning did not contain any text (e.g. a tweet that only contains a URL or an @mention). We remove such tweets from the dataset. The same text cleaning step was performed before generating features/embeddings for all classifiers described in the paper.



 



\subsection{Classifiers}
We tried, both, a traditional SVM classifier and the neural-networks (deep learning) based classifiers.

\subsubsection{SVM with TF-IDF features}
Support Vector Machine (SVM) is  a classifier of choice for many text classification tasks \cite{suykens1999least}. The classifier is fast to train and performs reasonably well on wide-range of tasks. As SVM takes on one set of features, we only use the reply text to train the model. The classifier takes TF-IDF (Term frequency- inverse document frequency) \cite{salton1988term} as input and predicts the two stance classes. We would expect that this simple model cannot effectively learn to compare the source and the reply text as is needed for good stance classification. However, we find that this model is still able to get reasonable accuracy and therefore serves as a good baseline. 

\subsubsection{LSTM  Model}

\begin{figure}[htb!]
    \centering
    \includegraphics[width=0.30\textwidth]{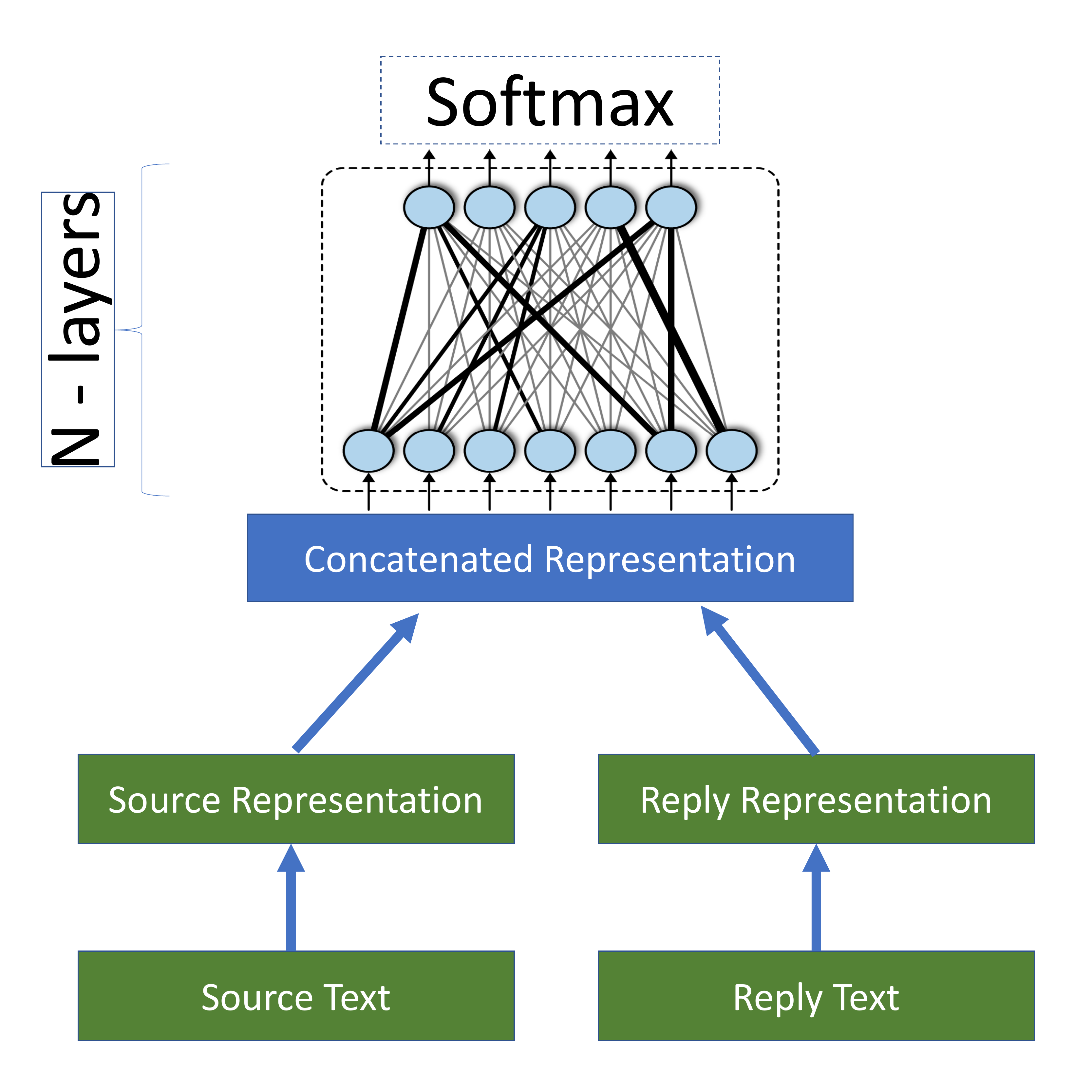}
    \caption{A model diagram for classifying conversations.}
    \label{fig:model_digram}
\end{figure}

As opposed to SVM text classifiers, a neural-network based models could be designed to use text-reply pair as input. One such model is shown in Fig. \ref{fig:model_digram}. A neural network based architecture that uses both source and reply can effectively compare target and reply posts and we  expect it to result in a better performance. The proposed model takes GLOVE \cite{pennington2014glove} Twitter based word embeddings (each with 200 size) as input, and contains a LSTM  layer (hidden state size = 200) that embeds text and reply to a fixed vector representations. A fully connected layer uses the last layer's output as input and a softmax layer on top to predict the final stance label. The fully connected layer is composed of relu activation unit followed by a dropout (20\%) and batch normalization. 


The model is trained using `RMSProp' optimizer using a categorical cross-entropy loss function. The learning rate (LR) we tried were in range $10^{-5}$ to $10^{-1}$. The  layers size we tried varied from $1$ to $3$. Once we find the best value for these hyper parameters by initial experiments on the Student Marches data by grid searching the parameter space and using F1-macro as criterion, the parameters remain unchanged during training and testing on other events. For the sequence based model, we find that a single fully connected layer and LR = $10^{-3}$ work the best. 



\begin{figure}[htb!]
    \centering
    \includegraphics[width=0.33\textwidth]{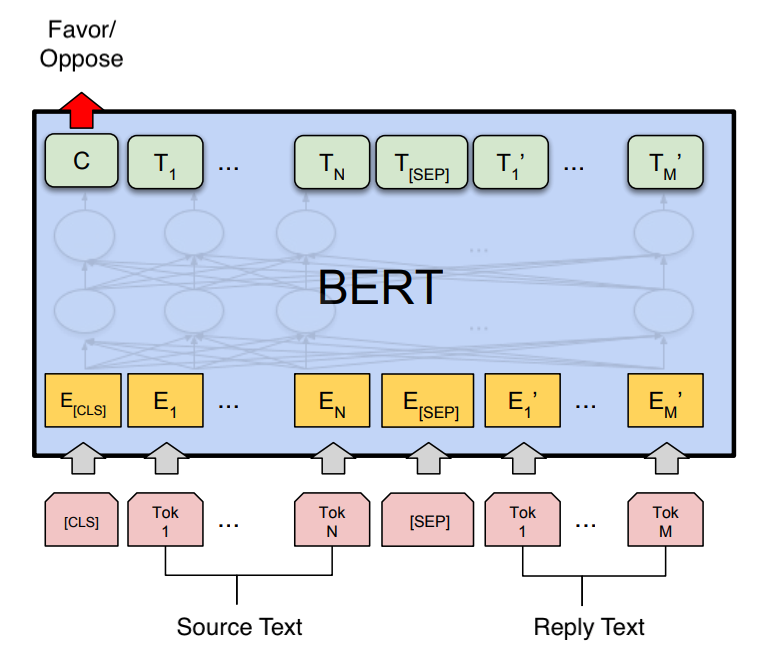}
    \caption{BERT Model of predicting stance in conversations as in favoring/opposing}
    \label{fig:bert}
\end{figure}

\begin{figure}[htb!]
    \centering
    \includegraphics[width=0.15\textwidth]{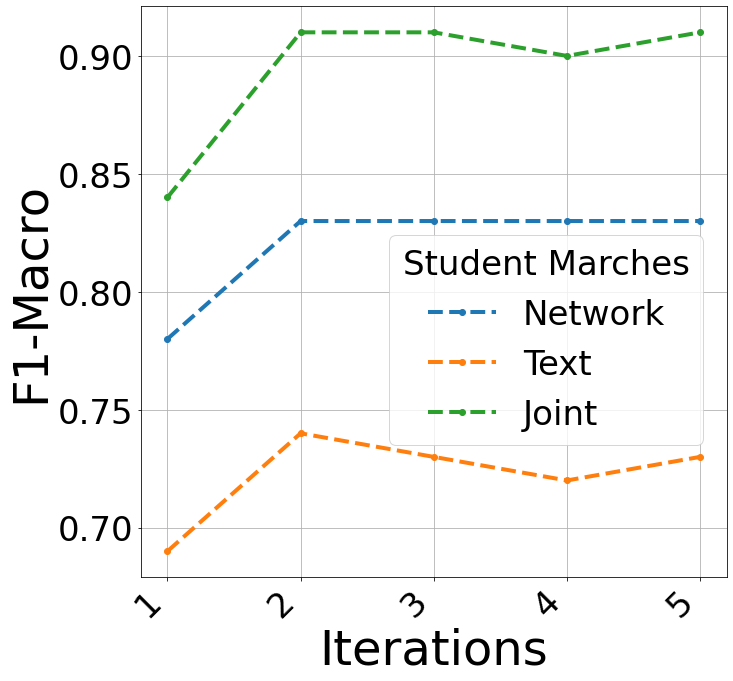}
\includegraphics[width=0.15\textwidth]{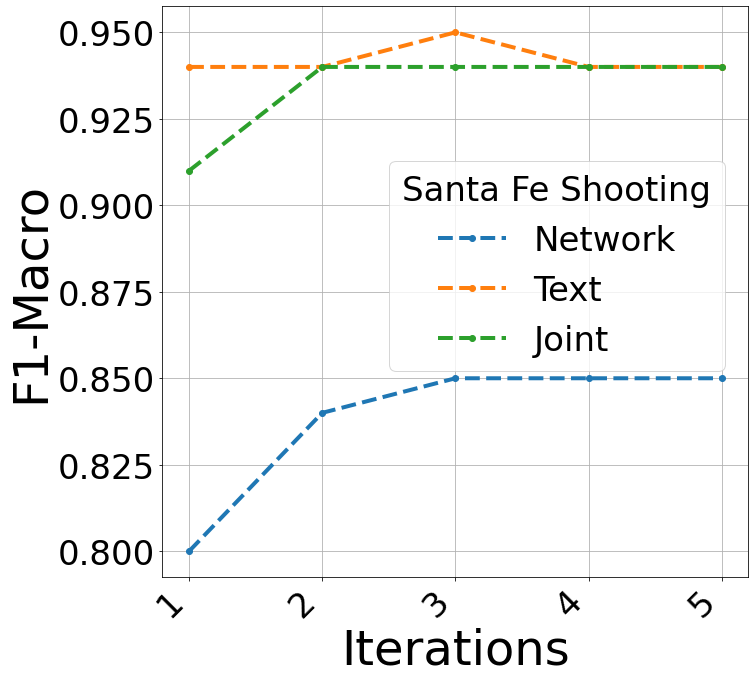}
\includegraphics[width=0.15\textwidth]{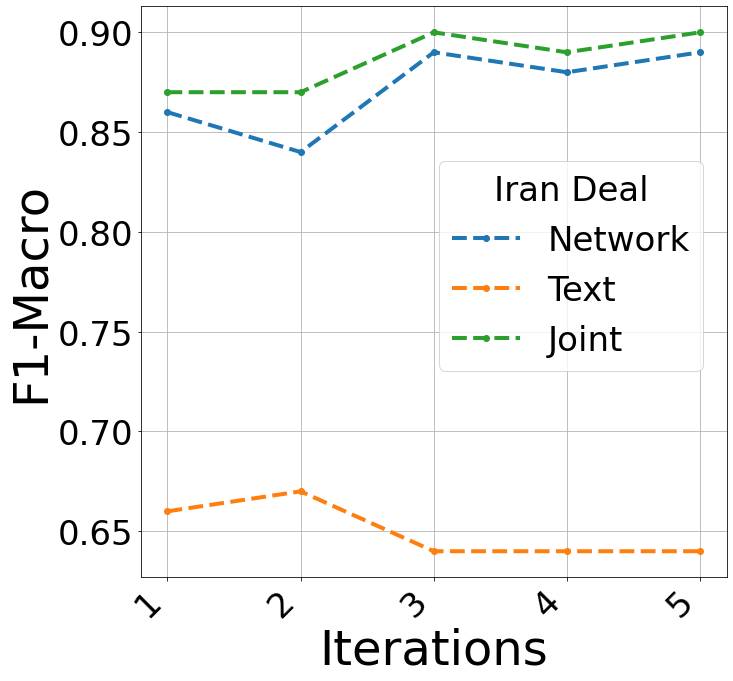}

    \caption{F1-Macro trend with co-training iterations for users' stance. }
    \vspace{-5mm}
    \label{fig:users_stance_accuracy}
\end{figure}

\subsubsection{BERT Model}
Attention based models have become very popular lately because of their superior performance on many tasks \cite{vaswani2017attention}. Models like BERT \cite{devlin2018bert} allow to achieve good performance by tuning the  pre-trained models. For the task of stance in conversations, we tried a BERT model with uncased tokenizer and Adam optimizer. A visual representation of a BERT model is shown in Fig. \ref{fig:bert}. As you can observe, the model takes source text, reply text, and a label for prediction. For tuning the model, we use learning rate of $3*10^{-5}$ , max sequence length =  $128$, and warm-up proportion $0.1$. For tuning, weight decay of $0.01$ is used for all layers except 'bias', 'LayerNorm.bias', 'LayerNorm.weight' for which weight decay is $0$.

\subsection{Training and Hyper Parameters Search}
As the dataset in not balanced, F1-Macro is used to compare the models as in many prior stance studies \cite{mohammad2017stance,lu2015biaswatch} . For co-training, we have five hyper parameters namely: 1) $\theta_u$, 2) $\theta^T$, 3) K (mixing parameter), 4) k (most used hashtags), and, 5) p (popular retweets count). The values of these parameters are determined by trials on one of the datasets (Student Marches). By a uniform grid search on SM dataset using F1-macro as criterion, we find the following values that work well: $k=250$, $p=1000$, $\theta_u = 0.7, \theta^T = 0.7,K = 0.2$. For $k$ and $p$, the parameter search range was 100 to 10,000. For $\theta_u$, $\theta^T$ and $K$, the search range was 0 to 1. Five iterations of co-training was used as the classifiers appear to converge after four iterations.

For training classifier with weak labels, a standard desktop was used for SVM, but for the neural-network based models, a machine with Nvidia GT-1080Ti GPU was used. Only weak labeled examples (excluding any test data) were used for training. The training of neural-networks took between half-an-hour to two hours but training SVM took less than half-an-hour. BERT, which has 110 million parameters, is much larger than LSTM which is in order of 200,000 parameters. However, for BERT we only tune the model whereas for LSTM we fully train the model (except the embeddings). 





\subsection{Results and Discussion}


We set two goals for the experiments, first to estimate users' stance classification performance and to verify if the co-training improves the performance over iterations, and second, to find if reply stance based models trained using weak-labels could perform equally well or better than supervised conversations stance classifiers.

For the first goal, we plot the performance of the users' stance classifiers with co-training iterations. As we can observe in the plot (see Fig. \ref{fig:users_stance_accuracy}), for all events, the joint models' performance has improved over the iterations. Based on F1-macro score, all three datasets get good performance of over 0.9 F1-macro. What is more interesting is how the performance of almost all models improved with iterations indicating co-training is useful. For text classifiers, for Student Marches and Iran Deal events, the F1-Macro score decreased after the second iteration which could be attributed to the fact that with every co-training iteration more data is added to the training set. If the new data is not as clean as the original (seed) data, the performance could decrease, and a pattern we commonly see in self-training methods \cite{nigam2000analyzing}. However, overall, the joint models improve over iterations showing the benefit of co-training.

\begin{center}
\begin{table}[!htbp]
\caption[Performance of conversations classifier on different dataset]{Performance of conversation classifiers on three datasets. Bold font indicates the best for a dataset.}
\small
\centering
\begin{tabular}{|p{1.9cm}|p{0.5cm}|p{0.5cm}|p{0.5cm}|p{0.5cm}|}
\hline

  Classifier Type $\downarrow$ Dataset $\rightarrow$& \multicolumn{1}{c|}{\begin{tabular}[c]{@{}c@{}}SM\\ (F1-Macro)\end{tabular}} & \multicolumn{1}{c|}{\begin{tabular}[c]{@{}c@{}}SS \\ (F1-Macro)\end{tabular}} & \multicolumn{1}{c|}{\begin{tabular}[c]{@{}c@{}}ID \\ (F1-Macro)\end{tabular}} & \multicolumn{1}{c|}{\begin{tabular}[c]{@{}c@{}}Mean\\ (F1-Macro)\end{tabular}} \\ 
\hline
\hline
 \multicolumn{5}{|l|}{\textbf{\textit{Baselines}}} \\ \hline
 Random &  0.50&0.46&0.52&0.49\\ \hline
 Majority & 0.40&0.42&0.41&0.41\\ \hline
 \hline
 \multicolumn{5}{|l|}{\textbf{\textit{Leave-one-out Event Based Supervised Classifiers}}} \\ \hline
 SVM using Reply Text & 0.52&0.56&0.56&0.55\\ \hline
LSTM using Source-Reply pair  & 0.59  & 0.59  & 0.56& 0.58\\ \hline
BERT using Source-Reply pair  & 0.61  & 0.63 & 0.59 & 0.61\\ \hline
 \hline
 \multicolumn{5}{|l|}{\textbf{\textit{Weakly Supervised Classifiers}}} \\ \hline

SVM using Reply Text & 0.58  & 0.64 & 0.41 & 0.54 \\ \hline
LSTM using Source-Reply pair  &   0.61 &  0.63  & 0.59 & 0.61 \\ \hline
BERT using Source-Reply pair  & \textbf{0.67} & \textbf{0.71}& \textbf{0.62} & \textbf{0.66}\\ \hline

\end{tabular}
\label{tbl:results_all_chapter_6}
\end{table}
\end{center}

For the second goal, we check and compare the performance of models trained using weak-labels and with leave-out-out event data. In leave-one-out event cross validation, out of three events, we train on two events data and test on the third event. We report the performance of models based on the test set. We summarize the performance of the models in Tab. \ref{tbl:results_all_chapter_6} in which we show the f1 score (macro) for all models for each dataset.

As we can observe, if we consider the mean values across events, the BERT model trained on weakly labeled data performs better than all others. SVM classifiers, though trained only on the replies, performs reasonably well, but its performance does not increase with weakly labeled data. In contrast, for the other two classifiers, using weakly labeled data helps to outperform the model trained on leave-one-out event data. The pattern in the results clearly shows that models trained with weak labels perform equally well or better.

\section{Analysis of COVID-19 Conversations}
\label{sec:covid}
To understand how useful our approach is beyond the experiments on human labeled dataset, we analyze Twitter discussions on COVID19. We collected COVID19 Twitter data in April of 2020 when topics `Open American Economy' and `Fire Dr. Fauci' were popular. In this analysis,  we try to answer `How different are the Pro and Anti users in their discussions?'. We extend this analysis in Appendix by answering two more questions:  1) What kind of media do these groups share?, and 2) Are the users primarily divided based on partisanship?

We describe the dataset highlighting the general statistics and the seed hashtags to get users' stance using our approach in Appendix. Once the stance of users are obtained, we can get the stance of  hashtags, mentions and URLs using the user-entity bipartite network $I$. Like we have mentioned earlier in methodology, using the network $I$, stance propagates from users to entities. If we assume $\theta_I$ to be a parameter that acts as a threshold for propagating stance from users to entities, the label propagation model could be written as:

\begin{equation}
    \tilde{S} \leftarrow \sigma'_{\theta_I}{(I' \cdot S^I)}
\end{equation}


where $\cdot$ is dot product, and $I'$ is the transpose of the matrix $I$. We use this formulation to estimate the stance given by hashtags and web URLs. Examining the results gives validation to the model and allows to answer the questions raised earlier.

\subsection{How different are the Pro/Anti groups?}
\begin{figure}[htb]
    \centering
\includegraphics[width=0.48\textwidth]{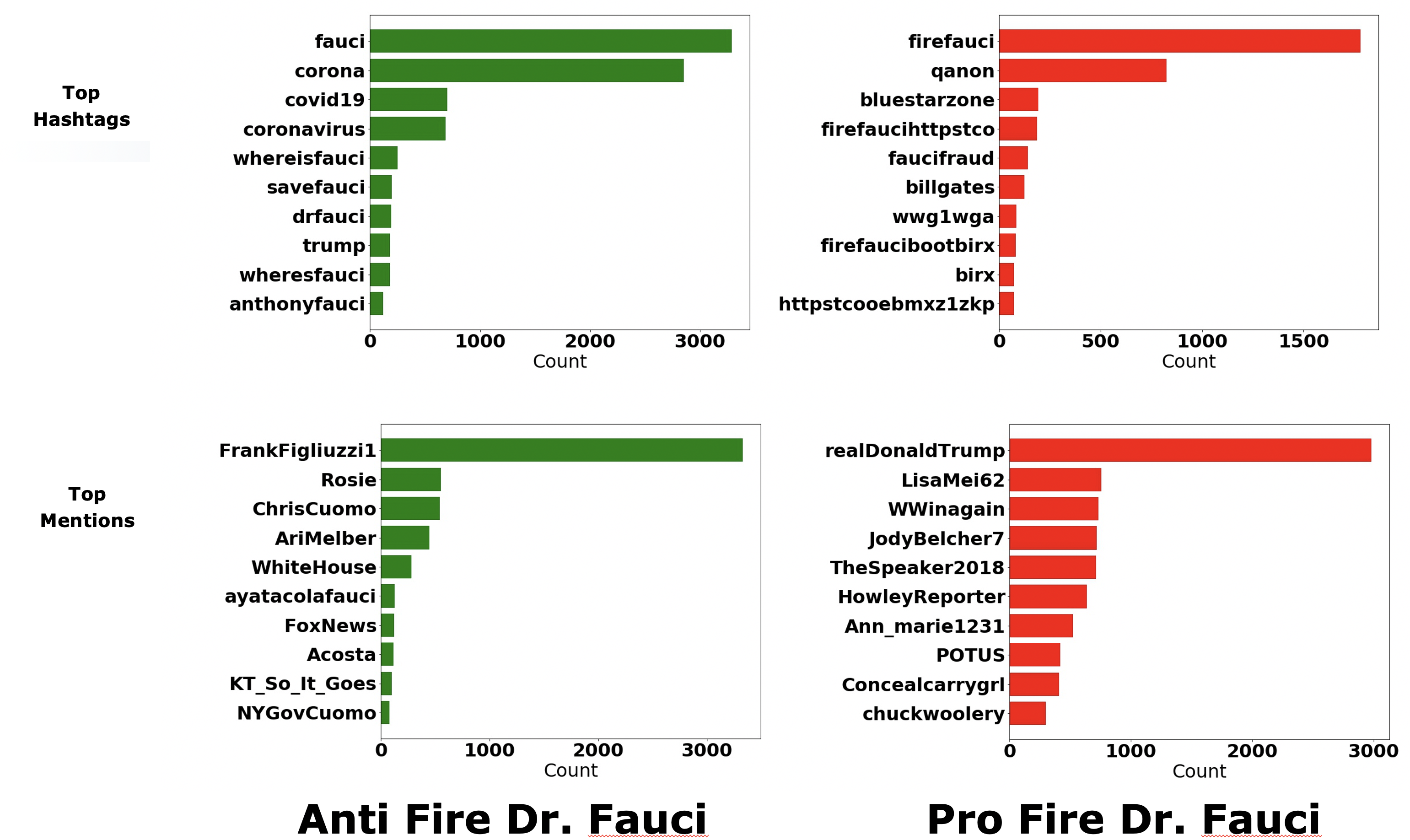}
    \caption{Fire Dr. Fauci: Comparison of top hashtags and top mentions used by the two groups discussing  }
    \label{fig:comparing_dr_fauci}
    \vspace{-3mm}
\end{figure}

\begin{figure}[htb]
    \centering
\includegraphics[width=0.48\textwidth]{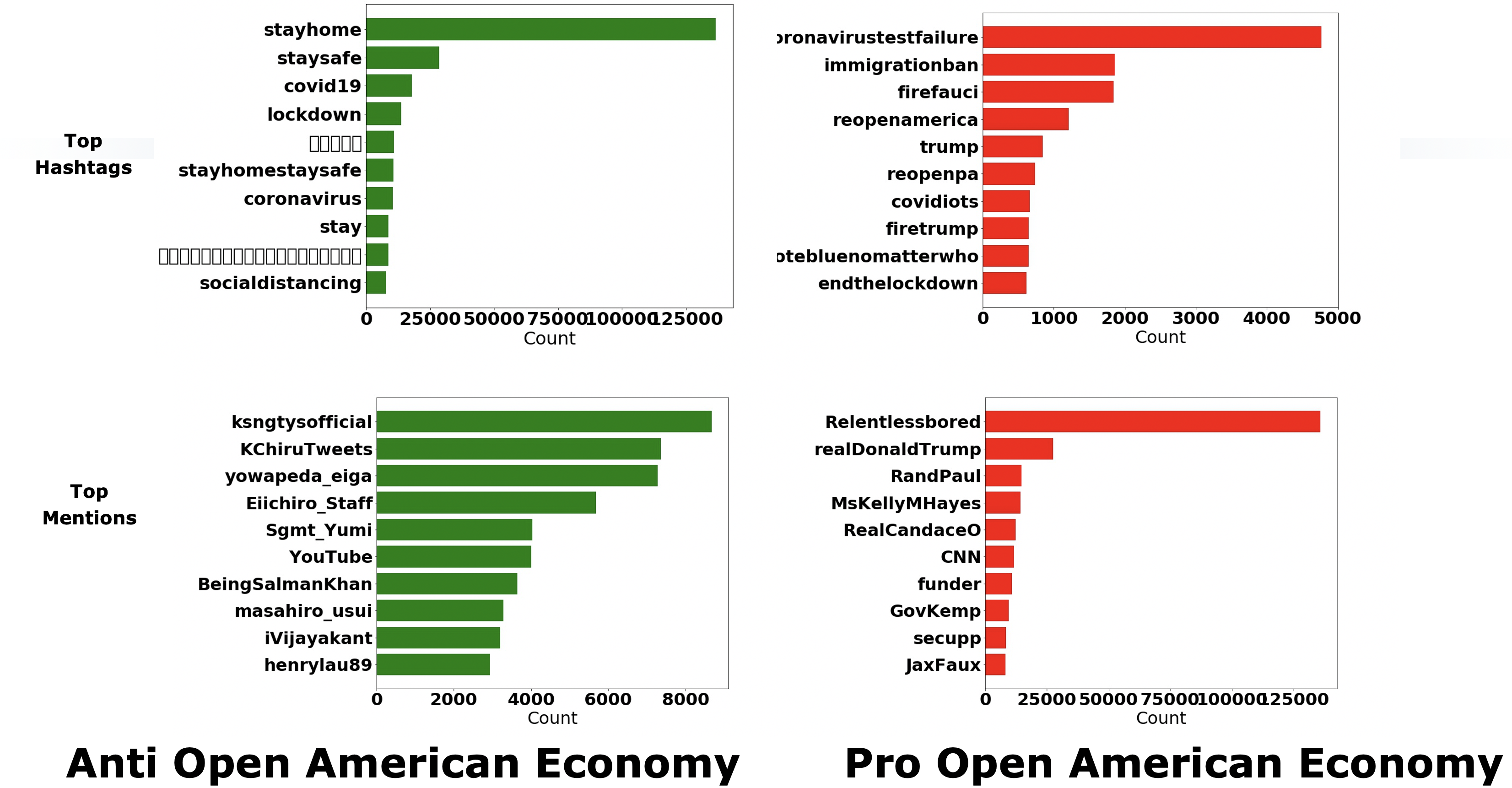}
    \caption{Open American Economy: Comparison of top hashtags and top mentions used by the two groups discussing `Open American Economy' }
    \label{fig:comparing_open_us}
    \vspace{-7mm}
\end{figure}

To understand the difference in the two groups (sides), we explore the top hashtags and mentions with pro and anti stance. We compare them using bar charts in Fig. \ref{fig:comparing_dr_fauci} and \ref{fig:comparing_open_us}.

As we can observe in Fig. \ref{fig:comparing_dr_fauci}, the top hashtags for the topic `Fire Dr. Fauci', on the anti side `savefauci',`wherisfacui'  are commonly used, and the top hashtags on the pro side include `firefauci' `faucifraud' etc. Intuitively, this result adds validity to our proposed approach. A similar difference is observed for the top mentions in the groups. As we can observe in Fig. \ref{fig:comparing_open_us} that compares pro and con discussions on opening the American economy, like `Fire Dr. Fauci', the the top hashtags and mentions differ a lot in terms of their usage.
\section{Related Work}
\label{sec:related_work}

\subsection{Stance in Conversations}
Stance in conversations was earlier explored as identifying stance in online debates or argumentation mining \cite{somasundaran2010recognizing}. Though stance-taking on social-media often mimic a debate, social-media posts are very short because of which many of the earlier developed methods don't directly transfer to conversations on platforms like Twitter. More recently,  stance in conversations on social media post is gaining popularity \cite{zubiaga2016stance,zubiaga2015crowdsourcing} with applications to predicting veracity of rumors. Unlike prior work, the focus of this work is using weak-supervision for stance in the context of Twitter replies. 

\subsection{Weak Supervision}
 Though weak labels are common in many areas of text mining, for opinion learning (as in stance of users in discussion) tasks, most work use labeled datasets \cite{mohammad2017stance,augenstein2016usfd,constance}. Mohammad et al. \cite{mohammad2017stance} briefly discussed the idea of using unlabeled tweets for improving their classification results. Misra et al. used a set of hashtags to build a topic-specific training corpus \cite{misra2016nlds}.   In our work,  we also to use weak-labels, but unlike prior work, we are interested in weak supervision for conversations. 
\subsection{Signed Link Prediction}
Signed link prediction has remained an active area of research for a while. Leskovec et al. \cite{leskovec2010signed} used thee data platforms (epinions, slashdot and Wikipedia) to show that the classical theory of structural balance tends common patterns of interaction, and extended the theory to directed graphs. In this and other work \cite{wang2017signed} on signed-link prediction,  ground-truth for links was known. In contrast, in this research, we try to predict stance in conversations (similar to signed links) without any labeled data for training. 

\section{Conclusion and Future Work}
\label{sec:conclusion}
In this research, we proposed a novel approach for learning stance in Twitter conversations. Our proposed method uses a few labeled hashtags to generate weak labels for the stance taken in conversations. We show the benefit of our approach by analyzing three topics with human labeled examples and demonstrating that the weak supervision approach can train models that could be better than fully supervised models. We further validate our approach by analyzing  COVD19 conversations showing how groups with opposing stance differ in using hashtags, mentions and URLs.

Despite the progress shown in this work, predicting stances in conversations remains a challenging problem. We expect to achieve a better performance by including other ways of weak supervision, e.g., by using emojis in reply posts.

\bibliographystyle{acl_natbib.bst} 
\bibliography{main}

\newpage
\setcounter{secnumdepth}{0}
\section{Stance Dataset Collection Method}
\label{stc_methodology}
\begin{figure}[ht] 
    \centering
    \includegraphics[width=0.48\textwidth]{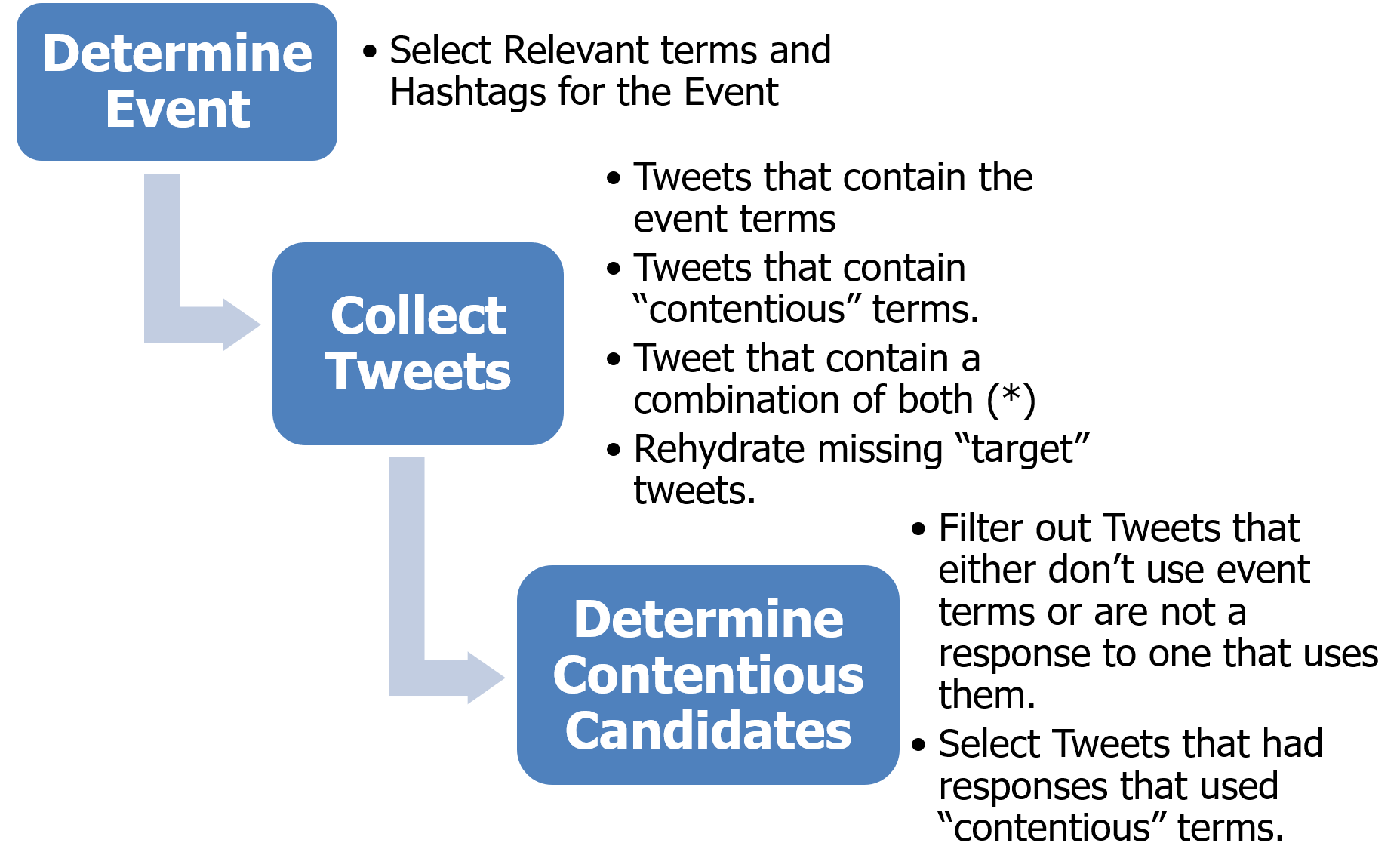}
     \caption{Methodology developed for the collection of contentious tweet candidates for a specific event.}
     \label{fig:col_method}
 \end{figure}

 Figure \ref{fig:col_method} summarizes the methodology developed to construct the datasets that skews towards more contentious conversation threads. We describe the steps in details next.
 
The first step requires finding the event related terms that could be used to collect the source (also called target) tweets. Additionally, as the focus is on getting more replies that are denying the source tweet, we use a set of contentious terms used to filter the responses made to the source tweets.

\subsection{Step 1: Determine Event}
The collection process centered on the following events.
\begin{itemize}
    \item \textbf{Student Marches}:  This event is based on the `March for Our Lives' student marches that occurred on the 24 of March of 2018 in the United States. Tweets were collected from March 24 to April 11 of 2018. \\The following terms were used as search queries: \#MarchForOurLives, \#GunControl, Gun Control, \#NRA, NRA, Second Amendment, \#SecondAmendment.
    
    \item \textbf{Iran Deal}: This event involves the prelude and aftermath of the United States announcement of its withdrawal from the Joint Comprehensive Plan of Action (JCPOA), also known as the "Iran nuclear deal" on May 8, 2018. Tweets were collected from April 15 to May 18 of 2018. \\
    The following terms were used as search queries: Iran, \#Iran, \#IranDeal, \#IranNuclearDeal, \#IranianNuclearDeal, \#CancelIranDeal, \#EndIranNuclearDeal, \#EndIranDeal.
    
    \item \textbf{Santa Fe Shooting}: This event involves the prelude and aftermath of the Santa Fe School shooting that took place in Santa Fe, Texas, USA in May 18, 2018. \\
    Tweets were collected from May 18 to May 29 of 2018. For this event, the following terms were used as search queries: Gun Control, \#GunControl, Second Amendment, \#SecondAmendment, NRA, \#NRA, School Shooting, Santa Fe shooting, Texas school shooting.
    
    
\end{itemize}

The set of contentious terms used across all events are divided in 3 groups: hashtags, terms and fact-checking domains:
\begin{itemize}
    \item \textbf{Hashtags}: \#bogus,  \#deception, \#wrong,  \#gaslight, \#fakeclaim,\#gaslighting, \#hoax, \#disinformation, \#FakeNews.
    \item \textbf{Terms}:  bull**t, bs, false, lying, fake, there is no, lie, lies, wrong, there are no, untruthful, fallacious, disinformation, made up, unfounded, insincere, doesnt exist, misrepresenting, misrepresent, unverified, not true, debunked, deceiving, deceitful, unreliable, misinformed, doesn't exist, liar, unmasked, fabricated, inaccurate, gaslight, incorrect, misleading, deception, bogus,  gaslighting, mistaken, mislead, phony, hoax, fiction, not exist, FakeNews.
    \item \textbf{Domains}: politifact, factcheck, opensecrets, snopes.
\end{itemize}

\subsection{Step 2: Collect Tweets}
Using Twitter's REST and the Streaming API we collected tweets that used either the event or contentious terms (as described earlier). If the target of a response is not included in the collection, we obtained it from Twitter using their API. 


\subsection{Step 3: Determine Annotation Candidates}
A target-response pair is selected as potential candidate to label if the target contains any of the listed event terms and the response contains any of the contentious terms. If urls are in the tweet, they are matched at the domain level by using the \textit{urllib} library in Python. For  `General Terms' event collected pairs based solely on the responses regardless of the terms used in the target.

To reduce the sample size, we filtered the tweets on some additional conditions. We only used the responses that were identified by Twitter to be in English and excluded responses from a user to herself (as this are used to form threads). In order to simplify the labeling context, we also excluded responses that included videos, or that had targets that included videos and limited our sample set to responses to original tweets. This effectively limits the dataset to the first level of the conversation tree.




\section{COVID Dataset Statistics}
We collected Twitter data using the Twitter streaming API in month of April, 2020. The search terms used to collect data included popular COVID related terms like `covid', `covod19', `corona' etc. Table \ref{tbl:covid_stats} summarizes the dataset. After collecting the dataset, we first filtered the entire data based on topics using terms relevant to the topic. We then used seed hashtags (see Table \ref{tbl:covid_stats}) and the joint result of the co-training method  to find the users that are `Pro' and `Anti' on the topics. 



\begin{center}

\begin{table}[htb]
\caption{Seed Hashtag and Stance Statistics}
\footnotesize
 \begin{tabular}{|p{1.1cm}|p{1.7cm}|p{1.7cm}|p{0.99cm}|p{0.99cm}|} 
 \hline
 Topic  &  Pro Seed Tags & Anti Seed Tags & Pro Users (Tweets) Count & Anti Users (Tweets) Count\\ [0.5ex] 
 \hline
 \hline
Open American Economy  &  'reopenamericanow', 'endthelockdown' & 'stayhomestaysafe', 'stayhome'  & 601,536 (1,065,771) & 390,719 (1,631,422)\\
 \hline
Fire Dr. Fauci & 'firefauci', 'firedrfauci', 'faucithefraud', 
'savefauci', & 'fauciisahero', 'keepfauci', 'firetrumpkeepfauci'  &  20,753 (45,125) & 8,527 (16,290)\\
 \hline
President Trump  &  'trump2020'  ,'bestpresidentever'
                    &  'firetrump', 'trumpliesamericansdie'  & 637,072 (1,216,041)& 432,575 (1,707,464)\\
 \hline
 \end{tabular}
 \label{tbl:covid_stats}
\end{table}
\end{center}

\section{COVID19 Dataset Extended Analysis}
Here we present additional analysis of COVID19 dataset. We first present how the usage of media differed across different groups, and then try to find is users are primarily divided on partisan ground.

\subsection{What kind of media do the different groups use?}
It is known that news agencies have their own biases which leads to some set of user reading and sharing a certain type of news. We wanted to check if this media preference is also reflected in how users with differing stance share websites.

\begin{figure}[htb]
    \centering
    \includegraphics[width=0.49\textwidth]{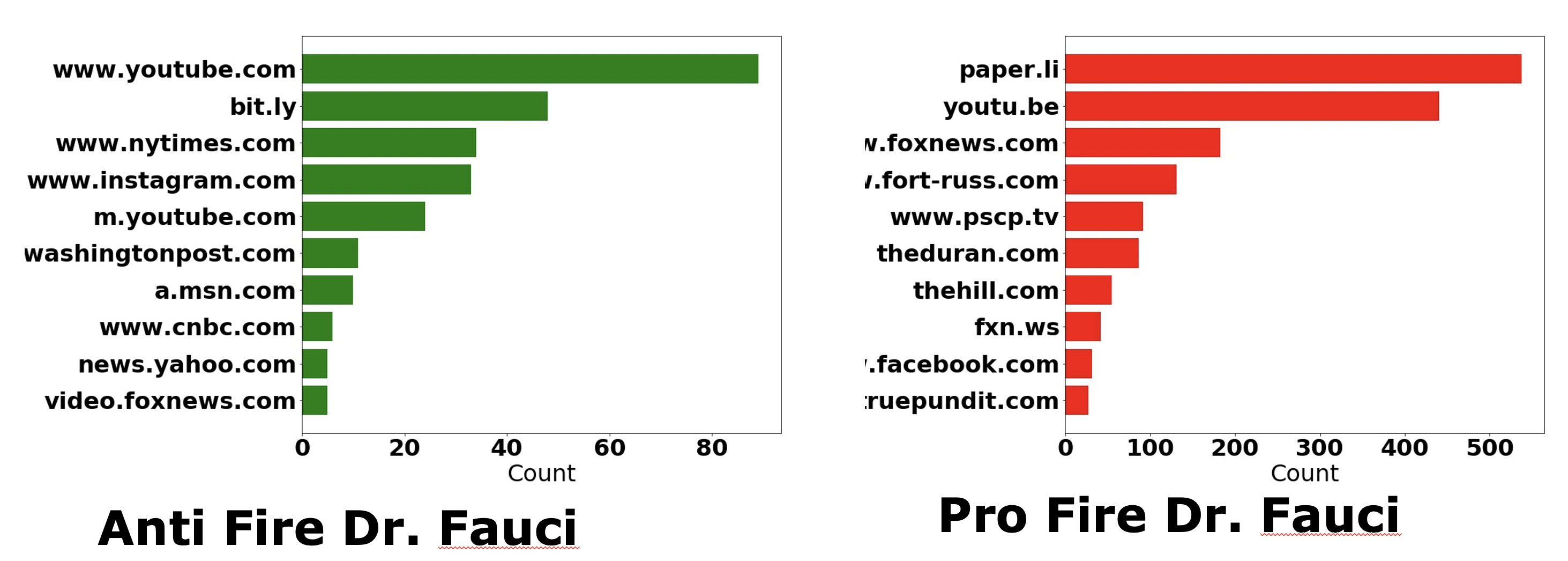}
    \caption{Firing Dr. Fauci: Comparison of top websites used by the two groups}
    \label{fig:websites_dr_fauci}
\end{figure}

\begin{figure}[htb]
    \centering
    \includegraphics[width=0.49\textwidth]{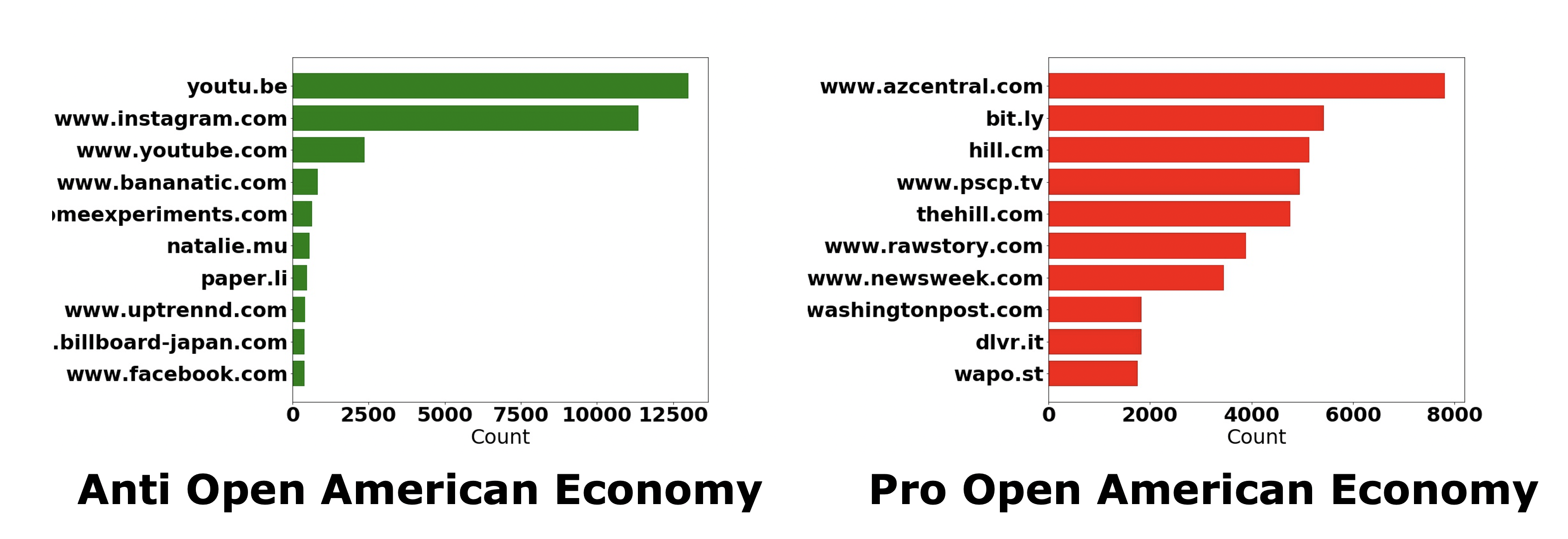}
    \caption{Open American Economy: Comparison of top websites used by the two groups discussing `Open American Economy' }
    \label{fig:websites_open_us}
\end{figure}

Figure \ref{fig:websites_dr_fauci} shows the web URL sharing pattern of users discussing `Fire Dr. Fauci' and Fig. \ref{fig:websites_open_us} shows the same pattern for users discussing the opening of the American economy. In both the plots, the stance driven preference is evident.

\subsection{Are the users primarily divided based on partisanship?}
\begin{figure}[htb]
    \centering
    \includegraphics[width=0.49\textwidth]{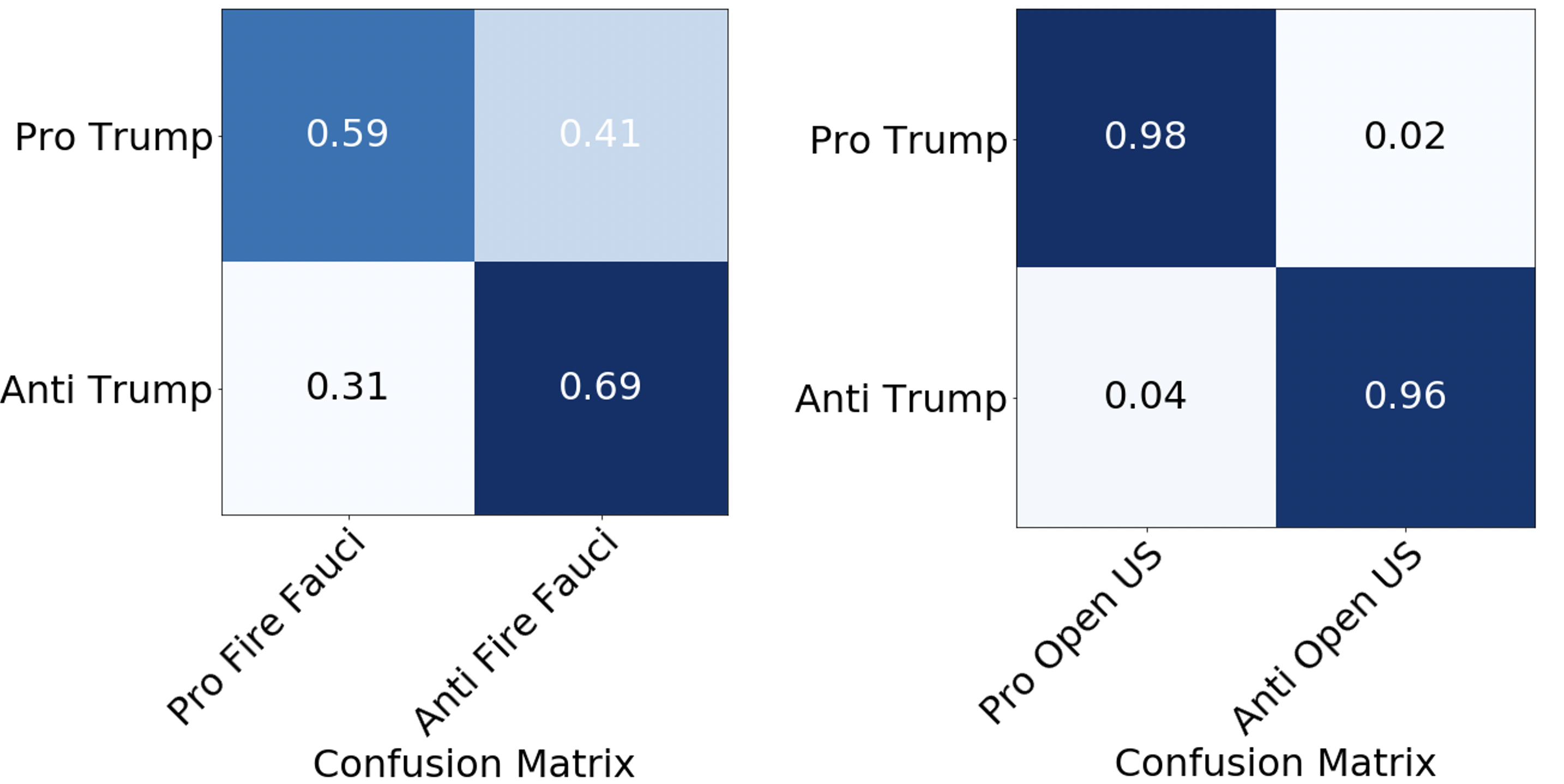}
    \caption{Comparison of divisiveness in topics based on partisanship}
    \label{fig:confusion_matrix}
\end{figure}

For understanding the effect of partisanship, we first estimated users stance about President Trump using seed hashtags mentioned in Tbl. \ref{tbl:covid_stats} and our proposed methodology. A stronger correlation between users that are `Pro Trump' and users in other topic would indicate that the discussion in the topics are primarily driven by partisanship. To visualize this correlation, we plot two confusion matrices in Fig. \ref{fig:confusion_matrix}. As we can observe, there is a high correlation between `Pro Trump' and `Open America' discussions, but the correlation between `Pro Trump' and `Pro Fire Dr. Fauci' is lower. This implies most users that support Trump also support the opening of the American economy, but many Trump supporters are against firing of Dr. Fauci. The overall analysis indicates that the various discussion on COVID-19 are not completely divided on partisan lines and even those users that favor Trump take different stance when it comes to policy decisions. 

To summarize, we used the methods developed for stance learning to analyze discussions on Corona virus. Besides allowing a high level picture of how users have varying stances on topics, our goal was to show that such analysis is possible for even large scale datasets with millions of users without requiring much labeling effort. The computational time taken for any of the above analysis was less than half an hour on a high end desktop proving that our approach is highly salable.

\end{document}